\documentclass[letterpaper]{article} 
\usepackage{aaai24}  
\usepackage{times}  
\usepackage{helvet}  
\usepackage{courier}  
\usepackage[hyphens]{url}  
\usepackage{graphicx} 
\usepackage{natbib}  
\usepackage{caption} 
\usepackage{algorithm}
\usepackage{algorithmic}

\usepackage{newfloat}
\usepackage{listings}
\DeclareCaptionStyle{ruled}{labelfont=normalfont,labelsep=colon,strut=off} 
\floatstyle{ruled}
\newfloat{listing}{tb}{lst}{}
\floatname{listing}{Listing}
\usepackage{amsmath}
\usepackage{amsfonts}
\usepackage{booktabs}

\title{H2G2-Net: A Hierarchical Heterogeneous Graph Generative Network Framework for Discovery of Multi-Modal Physiological Responses}
\author{
    Haidong Gu\textsuperscript{\rm 1},
    Nathan Gaw\textsuperscript{\rm 2},
    Yinan Wang\textsuperscript{\rm 3},
    Chancellor Johnstone\textsuperscript{\rm 2},
    Christine Beauchene\textsuperscript{\rm 4},\\
    Sophia Yuditskaya\textsuperscript{\rm 4},
    Hrishikesh Rao\textsuperscript{\rm 4},
    Chun-An Chou\textsuperscript{\rm 1}
}
\affiliations{
    \textsuperscript{\rm 1}Northeastern University, United States\\
    \textsuperscript{\rm 2}Air Force Institute of Technology, United States\\
    \textsuperscript{\rm 3}Rensselaer Polytechnic Institute, United States\\
    \textsuperscript{\rm 4}MIT Lincoln Laboratory, United States\\
    ch.chou@northeastern.edu
}

\usepackage{bibentry}

\begin{document}

\maketitle

\begin{abstract}
Discovering human cognitive and emotional states using multi-modal physiological signals draws attention across various research applications. Physiological responses of the human body are influenced by human cognition and commonly used to analyze cognitive states. From a network science perspective, the interactions of these heterogeneous physiological modalities in a graph structure may provide insightful information to support prediction of cognitive states. However, there is no clue to derive exact connectivity between heterogeneous modalities and there exists a hierarchical structure of sub-modalities. Existing graph neural networks are designed to learn on non-hierarchical homogeneous graphs with pre-defined graph structures; they failed to learn from hierarchical, multi-modal physiological data without a pre-defined graph structure. To this end, we propose a hierarchical heterogeneous graph generative network (H2G2-Net) that automatically learns a graph structure without domain knowledge, as well as a powerful representation on the hierarchical heterogeneous graph in an end-to-end fashion. We validate the proposed method on the CogPilot dataset that consists of multi-modal physiological signals. Extensive experiments demonstrate that our proposed method outperforms the state-of-the-art GNNs by 5\%-20\% in prediction accuracy.
\end{abstract}

\section{Introduction}

Cognition recognition is a rapidly growing research field that focuses on the development of data-driven and computational models to infer and interpret human cognitive states, such as attention, fatigue, perception, etc. Cognition recognition has numerous applications, such as health care, psychological assessment, education, and human-computer interaction \cite{gu2021detecting, gu2022optimizing, jia2021hetemotionnet, zheng2018emotionmeter, caballero4170114toward}. For instance, it can be used to analyze the cognitive states of pilots during flight practice, which provides insights on promptly adjusting and personalizing the teaching strategies to improve learning outcomes \cite{caballero4170114toward}.

Physiological data used in cognitive recognition studies usually include multiple heterogeneous modalities, such as electromyography (EMG), electrocardiography (ECG), photoplethysmography (PPG), pupil diameter (PD), and eye openness (EO), collected from different organ systems. Each modality may contain homogeneous sub-modalities. For example, the EMG modality includes sub-modalities of wrist flexor EMG and wrist extensor EMG in the unit $mV$. Hence, multi-modal physiological signals inherently have a hierarchical heterogeneous data structure. Furthermore, there also exist intra-modality and inter-modality interactions in the multi-modal physiological data that can objectively reflect human cognitive states.

In recent years, Graph neural networks (GNNs) have been widely applied to graph-structured data such as social networks \cite{chen2018fastgcn, hamilton2017inductive, wang2016structural}, citation networks \cite{kipf2016semi, velivckovic2017graph}, or traffic networks \cite{mo2021heterogeneous} and have achieved state-of-the-art results. GNN operates a convolution on an underlying graph structure by neighborhood messages passing and aggregation. However, a major limitation is that existing GNNs need to pre-define a graph structure that is usually homogeneous and non-hierarchical \cite{kipf2016semi, hamilton2017inductive, velivckovic2017graph}. For even more complex network problems including multi-modal physiological data, most existing GNNs fail to learn representative graphs effectively. The physiological network has multiple sub-graphs representing modalities (i.e., EMG, ECG, PD, etc.) and these sub-graphs contain multiple types of nodes. For example, EMG sub-graph includes two nodes of wrist flexor and extensor EMG in the unit $mV$, while PD sub-graph involves two nodes of left and right PD in the unit $mm$. In addition, there are no well-defined connections between modalities to form a physiological network. Therefore, the physiological network is called a hierarchical heterogeneous graph without a pre-defined graph structure.

The main \textbf{contributions} of our work are summarized as follows: (1) we propose a novel framework, H2G2-Net, for multi-modal data fusion that can automatically learn interactions and information flows among modalities, as well as model hierarchy and heterogeneity of modalities at the same time; (2) the generative graph of H2G2-Net is interpretable and provides valuable insights on interactions and information flows among modalities for prediction; and (3) we conduct extensive experiments on the benchmark CogPilot dataset and our H2G2-Net outperforms the state-of-the-art heterogeneous GNNs by a large margin of 5\%-20\% in classification accuracy of pilot training program difficulty levels.

\section{Dataset}

\begin{figure}[t]
    \centering 
    \includegraphics[width=\columnwidth]{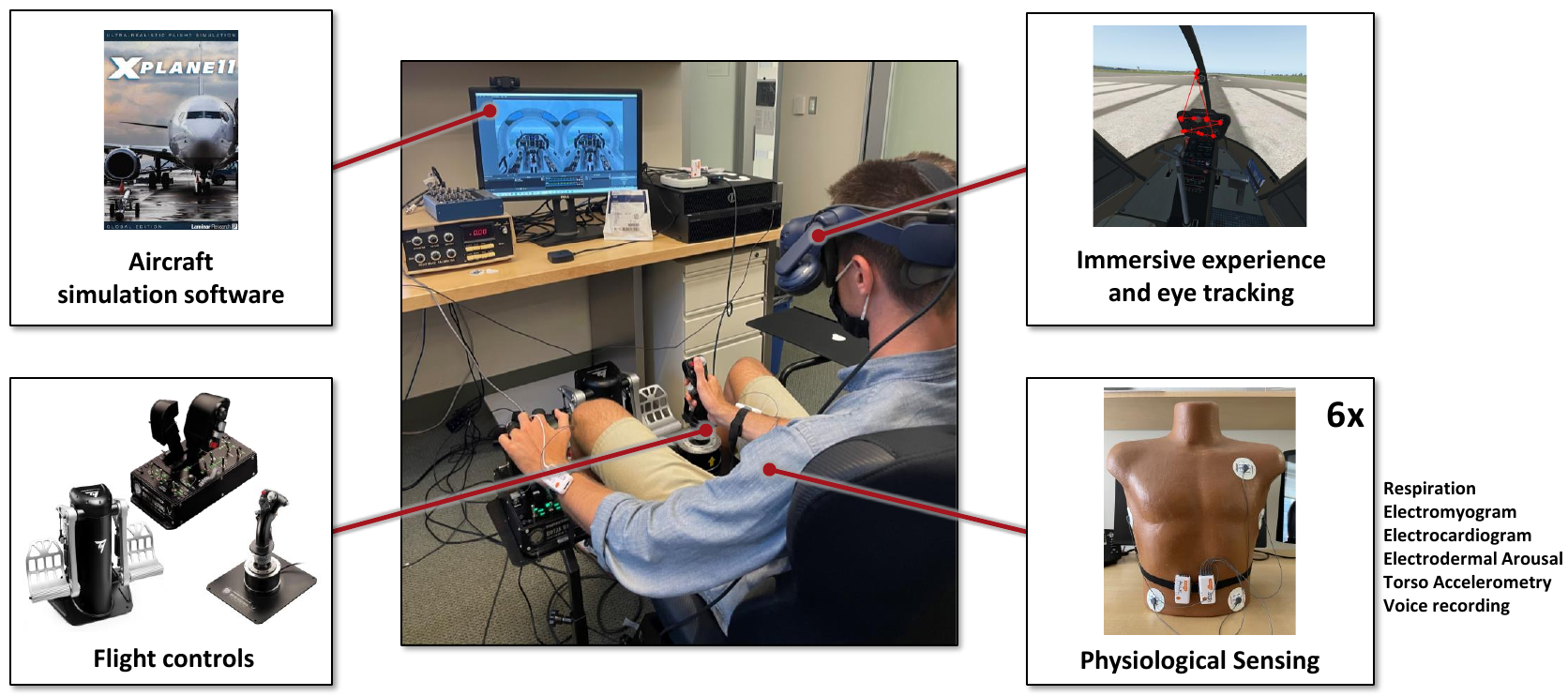}
    \caption{Data collection via physiological measurements recorded during immersive VR flight training simulations.}
    \label{fig_cogpilot}
\end{figure}

We collected the CogPilot dataset of multi-modal physiological recordings from 35 participants as they performed piloting tasks of varying difficulty in virtual reality as shown in Figure \ref{fig_cogpilot} \cite{raomultimodal}. In specific, the CogPilot dataset contains 9 modalities: (1) electromyography (EMG): muscle activation corresponding to the flexion and extension of the volunteer’s right wrist (joystick control) in the unit $mV$; (2) photoplethysmography (PPG): changes in blood volume measured on the left middle finger in the unit $mV$; (3) electrodermal activity (EDA): changes in electrical conductance of the skin, measured on the volunteer’s left hand in the unit $kOhms$; (4) electrocardiography (ECG): cardiac electrical activity, measured with electrodes across the chest in the unit $mV$; (5) respiration (RES): electrical signal of respiration derived using impedance pneumography in the unit $mV$; (6) accelerometry (ACC): three-axis accelerometry, measured on the right forearm as well as on the torso in the unit $m/s^2$; (7) gaze direction (GD): three-axis normalized gaze direction of the left and right eyes in $[0,1]$; (8) pupil diameter (PD): diameter of the left and pupils in the unit $mm$; (9) eye openness (EO): values representing how open the left and right eyes are in $[0,1]$. All recorded signals were time-synchronized using the Lab Streaming Layer to support multi-modal analysis. The CogPilot dataset can be downloaded from PhysioNet\footnote{https://physionet.org/content/virtual-reality-piloting/1.0.0/} \cite{goldberger2000physiobank}.

\section{Method}

\begin{figure}[t]
    \centering
    \includegraphics[width=\columnwidth]{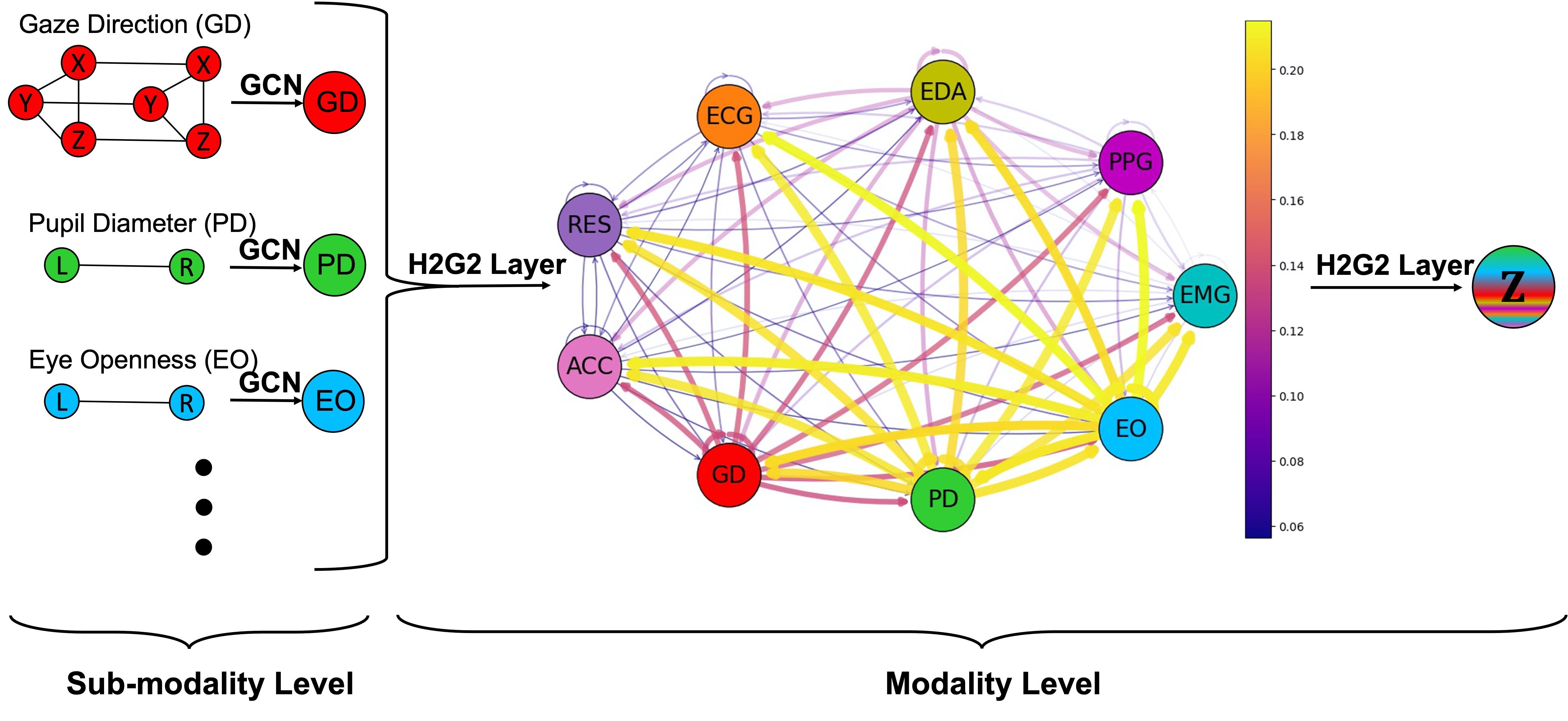}
    \caption{The architecture of the hierarchical heterogeneous graph generative network (H2G2-Net). Node denotes feature vector of modality or sub-modality. Node size measures amount of information. Node color represents modality. The feature vector of sub-modality is raw physiological signal. Edge denotes information flow among modalities or sub-modalities. Edge color and width indicate importance of information flow.}
    \label{fig_H2G2-Net}
\end{figure}

The hierarchical heterogeneous graph generative network (H2G2-Net) has two levels: modality and sub-modality, as shown in Figure \ref{fig_H2G2-Net}. The modality level operates on a heterogeneous graph defined by $(G, R)$, where $G=(g_1, g_2, \dots, g_m)$ is a set of subgraphs representing heterogeneous modalities, and $R=(r_{1,1}, r_{1,2}, \dots, r_{m,m})$ is a set of heterogeneous edges denoting relations among modalities. On the other hand, the sub-modality level operates on multiple homogeneous graphs that are each defined by a modality subgraph $g_i=(V, E)$, where $V=(v_1, v_2, \dots, v_n)$ is a set of homogeneous nodes representing channels of the same modality, and $E=(e_{1,1}, e_{1,2}, \dots, e_{n,n})$ is a set of homogeneous edges denoting interactions between channels. The H2G2-Net processes hierarchically from sub-modality level to modality level. So the hierarchical structure of multi-modal data can be well captured by H2G2-Net.

\subsection{Sub-modality level}

In the sub-modality level, the H2G2-Net applies GCNs \cite{kipf2016semi} to the multiple homogeneous graphs $g_i$ denoting modalities to learn graph representations of modalities. For modality $g_i=(V, E)$, let $X^{(l-1)}\in\mathbb{R}^{n\times d}$ be the input feature matrix of nodes in $V$ for the $l$-th GCN layer and $A\in\mathbb{R}^{n\times n}$ be the adjacency matrix defined by edge set $E$. The output node feature matrix is given as follows:
\begin{align}
    \label{sub-modality_level}
    X^{(l)} = \sigma \left( \hat{D}^{-\frac{1}{2}} \hat{A} \hat{D}^{-\frac{1}{2}} X^{(l-1)} \Theta \right),
\end{align}
where $\hat{A}=A+I\in\mathbb{R}^{n\times n}$ is the adjacency matrix $A$ with added self-loops, $\hat{D}$ is the diagonal degree matrix of $\hat{A}$ with $\hat{D}_{ii}=\sum_{j} \hat{A}_{i j}$, $\Theta\in\mathbb{R}^{d\times d}$ is the parameter matrix, and $\sigma$ is an activation function. Assume we set the number of GCN layers as $l$, then the graph representation vector $\mathbf{h}\in\mathbb{R}^{d}$ of modality $g_i$ is computed by row summation of node representation matrix $X^{(l)}\in\mathbb{R}^{n\times d}$ as follows:
\begin{align}
    \mathbf{h} = \sum_{i=1}^{n} X_i^{(l)}.
\end{align}
After the sub-modality level, we obtain feature vector $\mathbf{h}\in\mathbb{R}^{d}$ of each modality. The modality feature vectors are stacked into a matrix $H\in\mathbb{R}^{m\times d}$ as the input to the modality level.

\subsection{Modality level}

In the modality level, the H2G2 layers learn a sequence of weighted adjacency matrices that denotes a series of dynamic graph structures. The learned dynamic graph structures show information flows among modalities and identify useful meta-paths. For the physiological network $\mathbf{G}=(G, R)$, the representation matrix $H\in\mathbb{R}^{m\times d}$ of all modalities in $G$ is learned in the previous sub-modality level, but the graph structure defined by the edge set $R$ is not given. Let $H^{(l-1)}\in\mathbb{R}^{m\times d}$ be the input feature matrix of modalities in $G$ for the $l$-th H2G2 layer and $H^{(0)}=H$, then the operation of the $l$-th H2G2 layer on graph $\mathbf{G}$ without pre-defined graph structure is defined as
\begin{align}
    \label{modality_level}
    A &= softmax(\Phi) \\
    H^{(l)} &= \sigma \left( A H^{(l-1)} \Theta \right),
\end{align}
where $A\in\mathbb{R}^{m\times m}$ is the learned adjacency matrix that assigns weights to edges (i.e., learns significance of interactions), $\Phi\in\mathbb{R}^{m\times m}$ and $\Theta\in\mathbb{R}^{d\times d}$ are trainable weight matrices, $softmax$ denotes the row-wise softmax function, and $\sigma$ is an activation function. We can see in Equation (\ref{modality_level}) that each H2G2 layer learns a new adjacency matrix and generates a new graph structure. So by stacking H2G2 layers together, the H2G2-Net can learn a series of dynamic graph structures. The learned dynamic series of graph structures provides valuable insights on information flows among modalities and identifies significant meta-paths for prediction. The number of H2G2 layers can be considered as the meta-path length. At each H2G2 layer, every modalities updates its representation by aggregating messages passed from other modalities and the importance of messages are learned by the H2G2 layer as shown in weighted adjacency matrix. With the increasing of H2G2 layers, the information of each modality improves rapidly then saturates. Let the number of H2G2 layers be $l$, then the representation vector $\mathbf{z}\in\mathbb{R}^{d}$ of the physiological network $\mathbf{G}$ is calculated by row-wise summation of modality representation matrix $H^{(l)}\in\mathbb{R}^{m\times d}$ as follows:
\begin{align}
    \mathbf{z} = \sum_{i=1}^{m} H_i^{(l)}.
\end{align}
The representation vector $\mathbf{z}$ fuses multi-modal signals and characterizes information flows among modalities. Finally, we feed $\mathbf{z}$ into two fully-connected layers followed by a softmax layer for physiological network classification.

\section{Experiments}

We conduct extensive experiments on the CogPilot dataset which consists of multi-modal physiological signals collected during pilot flight training. The flight training program has $4$ predefined difficulty levels to evoke and induce cognitive states of pilots. Difficulty levels are defined by environmental settings such as wind speed, cloudness (visibility), and turbulence. The objective is to predict the difficulty levels using the multi-modal physiological data collected during the landing process.

Among $35$ subjects in the CogPilot dataset, we found missing modalities in some subjects (e.g., subject \#3 does not have PD signals). After thorough investigation, we selected $20$ subjects with complete modalities. Since the original physiological signals have high frequencies and different lengths, we downsample them all into $100$ data points.

To evaluate the effectiveness of representation learned by the H2G2-Net on multi-modal data, we compare it with the state-of-the-art heterogeneous GNNs. The GNN baselines include RGAT \cite{busbridge2019relational}, RGCN \cite{schlichtkrull2018modeling}, FiLM \cite{brockschmidt2020gnn}, HGT \cite{hu2020heterogeneous}, HAN \cite{wang2019heterogeneous}, GTN \cite{yun2019graph} and HetEmotionNet \cite{jia2021hetemotionnet}. Since all the above GNNs require a pre-defined graph structure, we construct the graph of multi-modal physiological data by making connections between different modalities. Then the baseline GNNs can operate on the manually defined graph.

We evaluate the model performance using a leave-one-subject-out (LOSO) cross validation approach. Data of all but one of the subjects are used for training and data of the remaining one subject is used for testing. This process is repeated until all subjects have been used for testing. The LOSO evaluation method ensures the model is tested on previously unseen subjects, so that it can provide a realistic assessment of model generalizability, i.e., the performance of model in real-world application.

\subsection{Results analysis}

We perform the task of classifying difficulty levels 1, 2 vs 3, 4 on the CogPilot dataset. Table \ref{tab_state} shows the average classification accuracy and standard deviation for the H2G2-Net and state-of-the-art GNNs. Our H2G2-Net achieves the best performance on the CogPilot dataset.

RGAT and RGCN perform the worst because they are, respectively, simple extensions of GAT and GCN to the heterogeneous graph. FiLM utilizes a feature-wise linear modulation to capture interactions between heterogeneous nodes. Hence, FiLM can extract more heterogeneous information and achieve better performance than GAT and GCN. HGT exploits the self-attention of the transformer architecture to learn relations on heterogeneous graphs and outperforms FiLM. Furthermore, HAN transforms the heterogeneous graph into homogeneous graphs defined by manually selected meta-paths and then applies GAT on the transformed homogeneous graph. HAN utilizes meta-paths and achieves better performance than previous models that do not consider meta-paths. HetEmotionNet is a hybrid neural network that consists of a GTN for learning the meta-paths, a GCN for modeling the interactions, and a GRU for capturing the temporal and spectral dependency. Although HetEmotionNet achieves relatively good performance, it requires pre-defined graph structure and does not consider the hierarchical information of multi-modal data like other baselines. In contrast, our H2G2-Net can automatically learn the graph structure and takes the hierarchical information of multi-modal data into consideration.

Therefore, our H2G2-Net can adequately learn comprehensive information and achieves the highest accuracy of $79\%$. Meanwhile, the H2G2-Net achieves the lowest standard deviation of $11\%$, which demonstrates that the H2G2-Net is very stable.

\begin{table}[t]
    \centering
    \small
    \begin{tabular}{cccc}
        \toprule
        Method & Mean Accuracy & Std Accuracy \\
        \midrule
        RGAT \shortcite{busbridge2019relational} & 60\% & 13\% \\
        RGCN \shortcite{schlichtkrull2018modeling} & 61\% & 11\% \\
        FiLM \shortcite{brockschmidt2020gnn} & 63\% & 17\% \\
        HGT \shortcite{hu2020heterogeneous} & 67\% & 12\% \\
        HAN \shortcite{wang2019heterogeneous} & 69\% & 12\% \\
        HetEmotionNet \shortcite{jia2021hetemotionnet} & 74\% & 16\% \\
        \midrule
        \textbf{H2G2-Net (2023)} & \textbf{79\%} & \textbf{11\%} \\
        \bottomrule
    \end{tabular}
    \caption{Results on the classification task.}
    \label{tab_state}
\end{table}

\subsection{Ablation studies}

To verify the effectiveness of multi-modal data fusion, we perform ablation studies on the classification task of difficulty levels 1, 2 vs 3, 4. We design two variants of H2G2-Net, which are described as follows: (1) H2G2-Net (NoETK): this variant removes three eye tracking (ETK) modalities PD, EO and GD from the multi-modal data to verify the effectiveness of using eye tracking modalities; and (2) H2G2-Net (ETK): this variant removes all the modalities except three ETK modalities from the multi-modal data to verify the effectiveness of other modalities excluding three ETK modalities.

The accuracy of H2G2-Net (ETK) is $72\%$ which is larger than H2G2-Net (NoETK)'s accuracy of $59\%$. This result indicates that the three eye tracking modalities PD, EO and GD are more effective than other modalities. Because eye tracking signals are more sensitive to cognitive states. Furthermore, H2G2-Net achieves the highest accuracy of $79\%$ and outperforms H2G2-Net (NoETK) and H2G2-Net (ETK) by a large margin (10\%-20\%), which demonstrates that fusing multi-modal data further improves the performance.

\subsection{Interpretability of the H2G2-Net}

\begin{figure}[t]
    \centering
    \includegraphics[width=\columnwidth]{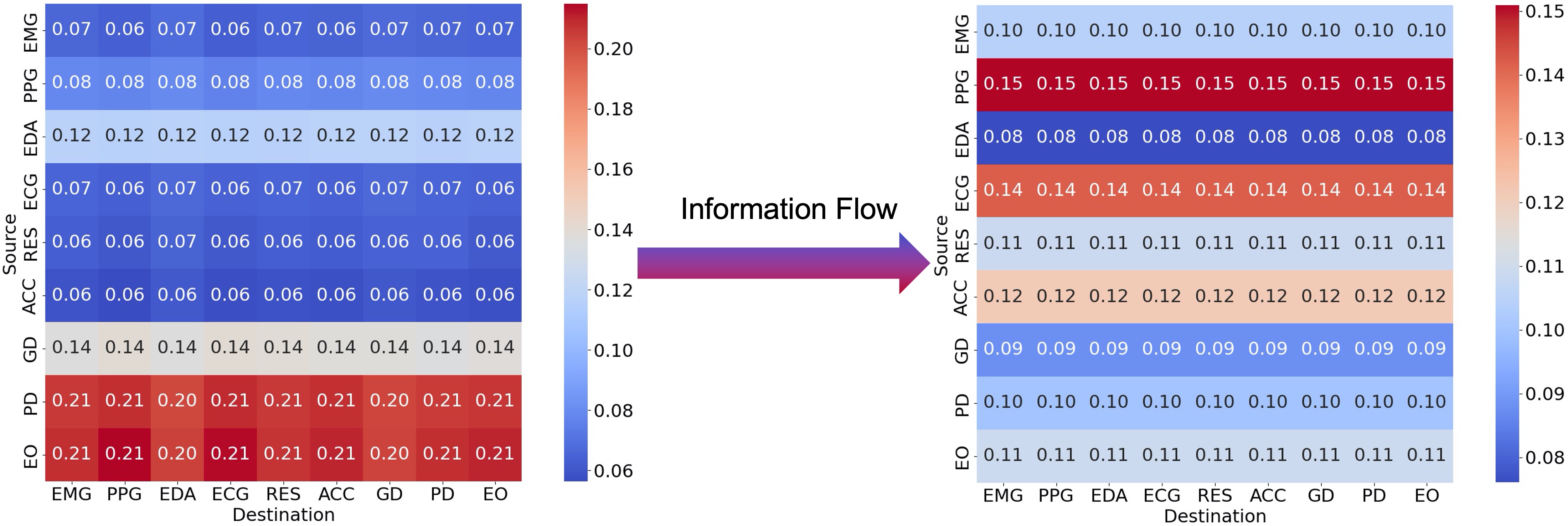}
    \caption{The sequence of adjacency matrices (i.e., dynamic graph structures) learned by the H2G2-Net.}
    \label{fig_meta}
\end{figure}

For the CogPilot dataset without a pre-defined graph structure, our H2G2-Net learns a sequence of adjacency matrices as shown in Figure \ref{fig_meta}. The series of adjacency matrices represents the dynamic graph structures. In specific, the entries of the adjacency matrices measure the importance of information flows among modalities. For example, in the first adjacency matrix, the importance of information flows from EO and PD to other modalities are all around $0.21$ which are the highest. It makes sense that when pilots perform flight task, they first observe the environment using eyes and generate EO and PD signals. Next, the environment information is transmitted to brain to form the cognition of task difficulty. Then, the cognitive state influences physiological responses such as EMG, PPG, EDA, ECG, RES and ACC. What are the next important information flows after \{EO, PD\} $\rightarrow$ \{EMG, PPG, EDA, ECG, RES, ACC\}? In the second adjacency matrix, we discover the information flows from ECG and PPG to other modalities are most important. Therefore, the next important information flows after \{EO, PD\} $\rightarrow$ \{EMG, PPG, EDA, ECG, RES, ACC\} are \{ECG, PPG\} $\rightarrow$ \{EMG, EDA, RES, ACC, EO, PD\}.

From above interpretation, we can identify significant meta-paths (i.e., information flows) like \{EO\} $\rightarrow$ \{PPG\} $\rightarrow$ \{RES\}. It is explainable that when subjects are in a high attention state, a greater tendency of eye openness is captured in the EO signal, followed by a heart rate acceleration reflected in the PPG signal, and then an increase of breath rate in the RES signal. Hence, the meta-path \{EO\} $\rightarrow$ \{PPG\} $\rightarrow$ \{RES\} is significant for prediction of cognitive states. In this way, we can interpret the dynamic graph structures by identifying important meta-paths and information flows. We believe that the interpretability of H2G2-Net provides valuable insights into prediction tasks by learning the importance of information flows in the network.

\section{Conclusion}

In this paper, we propose a hierarchical heterogeneous graph generative network (H2G2-Net) for multi-modal data fusion of physiological signals. Unlike existing GNNs that require a pre-defined graph structure, the H2G2-Net can automatically learn the graph structure by optimizing the modality-level adjacency matrix. In addition, existing GNNs do not consider the hierarchical information of multi-modal data, while the H2G2-Net can model the hierarchy by its two-level architecture. Extensive experiments on the benchmark CogPilot dataset demonstrate that the H2G2-Net outperforms the state-of-the-art GNNs by a large margin. Furthermore, interpreting the dynamic graph structures learned by H2G2-Net provides valuable insights on the information flows in the multi-modal data. Since the H2G2-Net can be combined with existing GNNs like GCN in our implementation, we expect that our framework can provide new ways for GNNs to optimize graph structures by themselves resulting in more effective representation learning on graphs. In the future, we will investigate the effectiveness of H2G2-Net combined with other state-of-the-art GNNs rather than GCN.

\section*{Acknowledgments}

This research was sponsored by the United States Air Force Research Laboratory and the Department of the Air Force Artificial Intelligence Accelerator and was accomplished under Cooperative Agreement Number FA8750-19-2-1000. This research was also partially supported by the Air Force Office of Scientific Research (AFOSR) under the Dynamic Data and Information Processing (DDIP) portfolio via Grant 23AFCOR003. The views and conclusions contained in this document are those of the authors and should not be interpreted as representing the official policies, either expressed or implied, of the Department of the Air Force or the U.S. Government. The U.S. Government is authorized to reproduce and distribute reprints for Government purposes notwithstanding any copyright notation herein.

\bibliography{aaai24}

\end{document}